\documentclass[sigconf]{acmart}
\usepackage{subcaption}
\usepackage{multirow}
\usepackage{xcolor}

\AtBeginDocument{%
  }

\makeatletter
\def\@ACM@copyright@check@cc{}
\makeatother

\copyrightyear{2025}
\acmYear{2025}
\setcopyright{cc}
\setcctype{by}
\acmConference[FAccT '25]{The 2025 ACM Conference on Fairness,
Accountability, and Transparency}{June 23--26, 2025}{Athens, Greece}
\acmBooktitle{The 2025 ACM Conference on Fairness, Accountability, and
Transparency (FAccT '25), June 23--26, 2025, Athens,
Greece}\acmDOI{10.1145/3715275.3732066}
\acmISBN{979-8-4007-1482-5/2025/06}

\begin{document}
\renewcommand{\abstractname}{Abstract}
\title{Component-Based Fairness in Face Attribute Classification with Bayesian Network-informed Meta Learning}

\author{Yifan Liu}
\affiliation{%
  \department{School of Information Sciences}
  \institution{University of Illinois Urbana-Champaign}
  \city{Champaign}
  \state{Illinois}
  \country{USA}
}
\email{yifan40@illinois.edu}

\author{Ruichen Yao}
\affiliation{%
  \department{School of Information Sciences}
  \institution{University of Illinois Urbana-Champaign}
  \city{Champaign}
  \state{Illinois}
  \country{USA}
}
\email{ryao8@illinois.edu}

\author{Yaokun Liu}
\affiliation{%
  \department{School of Information Sciences}
  \institution{University of Illinois Urbana-Champaign}
  \city{Champaign}
  \state{Illinois}
  \country{USA}
}
\email{yaokunl2@illinois.edu}

\author{Ruohan Zong}
\affiliation{%
  \department{School of Information Sciences}
  \institution{University of Illinois Urbana-Champaign}
  \city{Champaign}
  \state{Illinois}
  \country{USA}
}
\email{rzong2@illinois.edu}

\author{Zelin Li}
\affiliation{%
  \department{School of Information Sciences}
  \institution{University of Illinois at Urbana-Champaign}
  \city{Champaign}
  \state{Illinois}
  \country{USA}}
\email{zelin3@illinois.edu}

\author{Yang Zhang}
\affiliation{%
  \department{School of Information Sciences}
  \institution{University of Illinois at Urbana-Champaign}
  \city{Champaign}
  \state{Illinois}
  \country{USA}}
\email{yzhangnd@illinois.edu}

\author{Dong Wang}
\affiliation{%
  \department{School of Information Sciences}
  \institution{University of Illinois at Urbana-Champaign}
  \city{Champaign}
  \state{Illinois}
  \country{USA}}
\email{dwang24@illinois.edu}

\renewcommand{\shortauthors}{Liu et al.}


\begin{abstract}
The widespread integration of face recognition technologies into various applications (e.g., access control and personalized advertising) necessitates a critical emphasis on fairness. While previous efforts have focused on demographic fairness, the fairness of individual biological face components remains unexplored. In this paper, we focus on face component fairness, a fairness notion defined by biological face features. 
To our best knowledge, our work is the first work to mitigate bias of face attribute prediction at the biological feature level. 
In this work, we identify two key challenges in optimizing face component fairness: attribute label scarcity and attribute inter-dependencies, both of which limit the effectiveness of bias mitigation from previous approaches. To address these issues, we propose \textbf{B}ayesian \textbf{N}etwork-informed \textbf{M}eta \textbf{R}eweighting (BNMR), which incorporates a Bayesian Network calibrator to guide an adaptive meta-learning-based sample reweighting process. During the training process of our approach, the Bayesian Network calibrator dynamically tracks model bias and encodes prior probabilities for face component attributes to overcome the above challenges.
To demonstrate the efficacy of our approach, we conduct extensive experiments on a large-scale real-world human face dataset. Our results show that BNMR is able to consistently outperform recent face bias mitigation baselines. Moreover, our results suggest a positive impact of face component fairness on the commonly considered demographic fairness (e.g., \textit{gender}). Our findings pave the way for new research avenues on face component fairness, suggesting that face component fairness could serve as a potential surrogate objective for demographic fairness. The code for our work is publicly available~\footnote{https://github.com/yliuaa/BNMR-FairCompFace.git}.
\end{abstract}
\maketitle

\section{Introduction}

In recent years, face recognition technologies powered by deep learning have become prevalent across various applications, ranging from automated customer service to access control and personalized advertising~\cite{parkhi_deep_2015}. While these technologies excel in prediction accuracy, recent studies have highlighted a significant disparity in performance across different demographic groups~\cite{pmlr-v80-kearns18a}. Bias in facial recognition can lead to identity verification failures, biased content recommendations, and inconsistent user experiences across different demographic groups~\cite{Robinson_2020_CVPR_Workshops}. Face attribute classification in face recognition is a key component in user profiling and personalization on online platforms~\cite{gong_attribute_2018,serna2019algorithmic}. Therefore, it is crucial to address algorithmic bias in face attribute classifiers, as these classifiers are central to ensuring fair predictions~\cite{howard_ugly_2018}.

To mitigate bias in machine learning powered classification systems, the majority of fairness-aware methods are based on different definitions of demographic fairness~\cite{dwork_fairness_2011,kusner_counterfactual_2018}, which fails to address discrimination at a \textbf{biological feature level}. For example, a demographically fair face attribute classifier can provide unbiased predictions for different genders but can still discriminate against faces with a certain nose or eyebrow shape.
Such discrimination against biological facial features can result in unequal access to services, such as facial recognition for secure logins. Marginalized groups may face higher rates of mis-identification or exclusion from digital platforms due to such discrimination.

In this paper, we define a new fairness notion grounded in the \textit{biological components} of facial features and examine this fairness principle within the context of face attribute classification tasks. Specifically, we define \textbf{face component fairness} as a composite group fairness notion that requires a classifier to be fair in terms of different biological components of a face (e.g., lips, eyebrows, nose). Following this definition, we ask the following research questions: 
\begin{itemize}
    \item \textbf{Can we train a face attribute classifier that is fair in terms of face component fairness?}
    \item \textbf{What is the relationship between face component fairness and demographic fairness (race, gender, age)?}
\end{itemize}

Our study examines algorithmic bias in face classification tasks by focusing on debiasing face component attributes, thereby enhancing the fairness of predictions in facial recognition systems. This work pivots the focus on fairness in machine learning from demographic fairness to the dimension of face component fairness.

Specifically, we consider group fairness notion that aims to achieve fair prediction across a set of subpopulations
featured by sensitive attributes following prior works~\cite{xu_fairgan_2018,kang_infofair_2022,shu_meta-weight-net_2019,learning-to-reweight-data,shui2022learning,yao2022improving}. For example, for sensitive attribute \textit{gender}, the subpopulations are people with different genders. Many of the existing algorithms on group fairness are restricted to only one subpopulation such as race or gender~\cite{shu_meta-weight-net_2019,ren_learning_2019,learning-to-reweight-data,lahoti_fairness_2020,chai2022fairness,shui2022learning, yao2022improving}. However, in ideal situations, ML models should make fair predictions for multiple potentially overlapping subpopulations. For example, a fair face attribute classifier is required to provide fair predictions for groups with different demographic attributes (e.g. race, gender, age) or with different biological features (e.g. hair color, skin color).
In our approach, we treat face component fairness as a form of group fairness, where each face component represents a distinct group.

\begin{figure}[t]
  \centering
  \includegraphics[width=\linewidth]{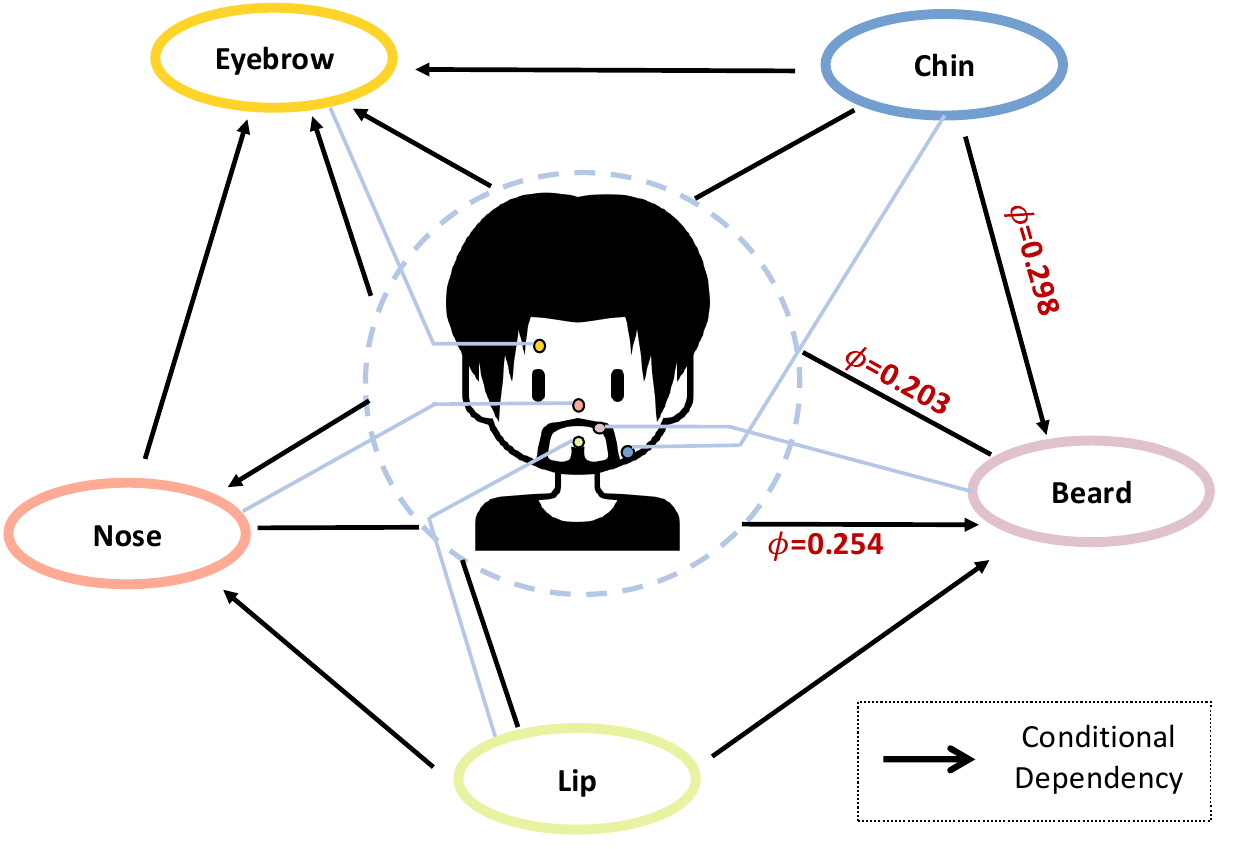}
  \Description{Illustration of face component dependency relations extracted by a Bayesian Network structural searching process in CelebA dataset.}
  \caption{Illustration of face component dependency relations extracted by a Bayesian Network structural searching process in CelebA dataset, where significant dependencies are highlighted with their $\phi$ values from chi-square test of independence. Face component fairness requires a classifier to be fair across various biological face components (e.g., lips, eyebrows, and nose). Achieving fairness at the component level requires unbiased predictions across correlated attributes, as biases in one component can affect others due to their inter-dependencies.}
  \label{fig:dependency}
\end{figure}

To mitigate algorithmic bias under face component fairness, we identify the following challenges: (1) 
There exist complex inter-dependencies between the attributes of each face component (see Figure~\ref{fig:dependency}), which makes it difficult to directly apply many existing debiasing algorithms developed based on the independence assumption of sensitive attributes.  Additionally, the inter-dependencies among face component attributes pose a non-trivial challenge in gradient-based fairness-aware optimization process, i.e., enhancing model performance for one face component attribute might introduce extra bias on another~\cite{pmlr-v80-kearns18a}. (2) The increasing number of face component attributes considered can incur extra computational overload. Prior work often accomplished unbiased fairness evaluation with a fairly sampled dataset in terms of all studied sensitive attributes~\cite{shu_meta-weight-net_2019,learning-to-reweight-data,ren_learning_2019}. However, due to the inter-dependencies among different face components, the number of possible combinations of sensitive attributes grows exponentially as each unique combination of attributes is treated as a distinct subpopulation. This makes it impractical to directly incorporate fairness metric evaluation into the training process, as it would result in excessive computational overhead.
Moreover, for face component fairness, such sampling process could introduce the attribute label scarcity problem and evaluation bias~\cite{yang_fairness_2020}. To directly incorporate optimization of group fairness into model training, fairness-aware optimization requires \textit{unbiased} and \textit{efficient} in-training fairness evaluation to guide the bias mitigation process. 

To address the challenges above, we propose \textbf{B}ayesian \textbf{N}etwork-informed \textbf{M}eta \textbf{R}e-weighting (BNMR), a novel algorithm that is designed for face component fairness with multiple face component attributes. We develop a meta-learning based algorithm enhanced with a Bayesian Network to model the complex dependency between biased model predictions and the face component attributes. Specifically, our algorithm learns to re-weight each data point based on sampled small fair validation set for each face component. Different from prior works~\cite{shu_meta-weight-net_2019,learning-to-reweight-data,ren_learning_2019}, our method designs a \textit{probabilistic graphical model} for conditional dependency modeling of face component attributes and develops a \textit{meta-learning} approach to adaptively reweighting input data, achieving a superior face component fairness-accuracy trade-off. In addition, our method does not require a balanced sampled dataset to perform as an exemplar of target distribution. Instead, we use a trained Bayesian Network to capture the prior distribution of the face component attributes, which enhances the scalability of our approach. We evaluate our method on a large-scale human face dataset with face component level annotations and demonstrate that our approach can effectively mitigate bias of different face component attributes. Furthermore, we observe that
reducing bias in face component attributes also interestingly improves demographic fairness. Our key contributions are summarized as follows: \begin{enumerate}
    \item We propose a novel fine-grained fairness notion called face component fairness, which is defined by biological components for face attribute classification tasks.
    \item We develop \textbf{B}ayesian \textbf{N}etwork-informed \textbf{M}eta \textbf{R}e-weighting (BNMR) to mitigate bias with respect to face component fairness. Our algorithm learns to dynamically re-weight samples in each batch to accommodate the gradient information from a calibrated fairness metric evaluation.
    \item We evaluate the BNMR on a large-scale human face dataset. Our results demonstrate that BNMR achieves the best overall performance in face component fairness across multiple face component attributes while maintaining comparable classification results.
    \item We investigate the relationship between face component fairness and demographic fairness (e.g., \textit{gender}) and find our method enhancing component-level fairness can also improve demographic fairness.
\end{enumerate}
\section{Related Work}
\subsection{Face Component Fairness}
In this paper, we define face component fairness to be a group fairness with multiple overlapping inter-dependent groups. Group fairness is a common fairness notion that aims to mitigate performance disparity between different subpopulations. This notion of fairness has been extensively studied in many domains such as credit scoring~\cite{feldman_certifying_2015} and healthcare~\cite{esteva_guide_2019}.
Prior works on group fairness typically aimed to ensure that task performance is independent or conditionally independent of attribute assignment.
 Such independence is further specified through metrics like equal opportunity, statistical parity~\cite{hardt_equality_2016}, and group sufficiency~\cite{shui2022learning}. Motivated by the underlying requirement, several prior works propose to mitigate this performance disparity by directly enforcing the independence as constraints during model training~\cite{cotter2019optimization,pmlr-v98-cotter19a}. 

Despite numerous studies have been conducted to address the attainment of demographic group fairness in face recognition tasks, our focus on face component fairness is a largely unexplored area of research. Methods focusing on fairness without demographics can be considered as a potential solution for face component fairness as these methods require no extra knowledge of group memberships~\cite{lahoti_fairness_2020,hashimoto_fairness_2018,chai2022fairness} and can be applied to different group fairness specifications. Shui et al.~\cite{shui2022learning} propose a probabilistic bi-level optimization method that considers approximating the fair distribution from multiple distributions. However, the bi-level optimization proposed can only address non-overlapping subgroups, which cannot be applied to ensure face component fairness that are interdependent. InfoFair~\cite{kang_infofair_2022} is a method that aims to achieve statistical parity~\cite{verma_fairness_2018} by enforcing the independence between the sensitive attributes and predictions via minimizing the mutual information. Different from prior works, our approach takes an explicit modeling of sensitive attributes with a Bayesian Network, thus is naturally scalable to multiple inter-dependent sensitive attributes. 

\subsection{Sample Reweighting for Bias Mitigation}
Our proposed approach follows sample reweighting, which is a commonly used paradigm to mitigate algorithmic bias and resolve distributional shift~\cite{learning-to-reweight-data}. For data-centric reweighting, earlier methods aim to mitigate bias based on ideas like importance sampling, sample correlation~\cite{kahn_methods_1953}, and rejection sampling~\cite{zadrozny_learning_2004}. More recent works aim to extract sample weights from class sizes and data difficulty~\cite{cui_class-balanced_2019,lin_focal_2018} . These algorithms typically consider static sample weights that are inferred from statistical estimation. While they provide intuitive mitigation of label bias, they do not directly consider bias in the training process. Recent work suggests a trained face classifier could still suffer from biased predictions with a fair sampled training set~\cite{wang_balanced_2019}, highlighting the need for a more flexible debiasing scheme in the realm of face recognition.

As an alternative, learning based re-weighting scheme is considered to be a more flexible debiasing scheme as it learns sample weights iteratively and directly from training process. Inspired by meta-learning algorithms~\cite{finn_model-agnostic_2017}, recent reweighting methods model the learning of sample weights following the ``learning-to-learn'' idea. The meta-learning formulation enables the learning of an adaptive sample weight function that can be modeled either implicitly or explicitly. For example, Dehghani et al.~\cite{dehghani_fidelity-weighted_2018} propose to use a Bayesian function approximator to model the weight function. Wu et al.~\cite{wu_learning_2018} use a neural network with attention mechanism to weight samples. Ren et al.~\cite{ren_learning_2019} propose a meta-learning algorithm that assigns weights to samples based on their gradient direction similarity to an unbiased meta gradient update. Albeit the simplicity of such implicit modeling, this method reduces the stability of weighting behavior and restricts the weight generalization. In contrast, meta-weight-net~\cite{shu_meta-weight-net_2019} learns an MLP with one hidden layer as weight function from a small amount of unbiased meta-data. FORML~\cite{learning-to-reweight-data} directly incorporates the fairness metric in the optimization process through sample reweighting. However, it still necessities a user-specified exemplar fairness validation dataset with fair batch samplings in the training process, which suffers from attribute label scarcity issue when considering multiple sensitive attributes. Our method differs from FORML as it does not require a fair sampling process, thus avoiding problems caused by data scarcity and excessive computation load. Compared to existing reweighting based bias mitigation methods for improving face component fairness~\cite{dehghani_fidelity-weighted_2018, learning-to-reweight-data,wu_learning_2018,ren_learning_2019,shu_meta-weight-net_2019}, our approach accounts for dependencies among sensitive face component attributes and predictions, resulting in a more accurate fairness evaluation and optimization.

\section{Methodology}
\begin{figure}[t]
  \centering
  \includegraphics[width=\linewidth]{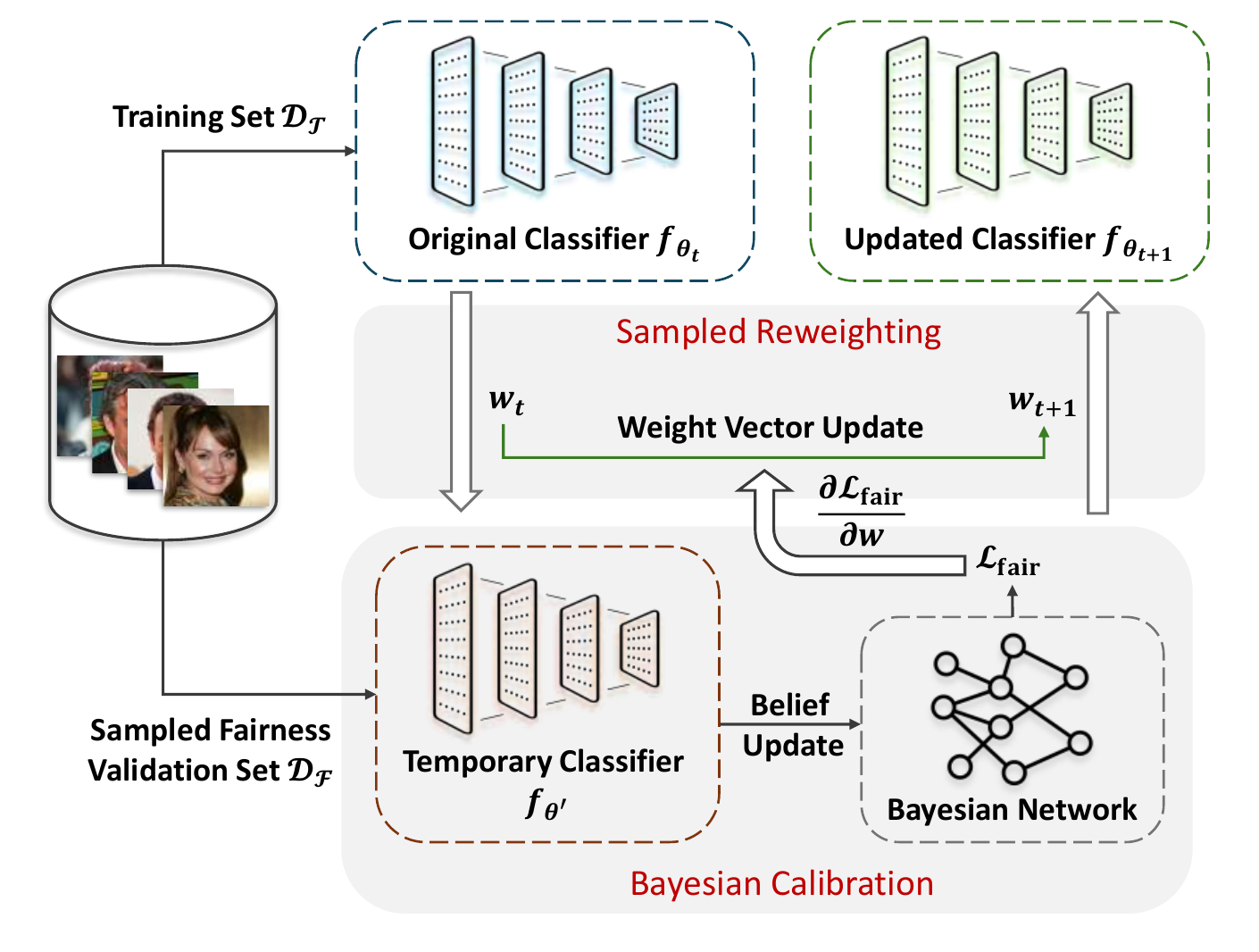}
  \caption{An overview of our method, which learns a weight vector $w$ with a Bayesian Network calibrator$(\Phi)$ in a fairness-aware meta-learning.}
  \label{fig:pipeline}
\end{figure}

In this section, we formally introduce our problem setting and present \textbf{B}ayesian \textbf{N}etwork-informed \textbf{M}eta \textbf{R}e-weighting (BNMR) method that aims to improve face component fairness via sample re-weighting. As illustrated in Figure~\ref{fig:pipeline}, our approach involves sample reweighting process and a Bayesian calibration process that facilitate effective and efficient fairness-aware training. 

\subsection{Face Component Fairness}\label{dependency}
We first define our proposed face component fairness. For a set of face component attributes, while it is possible to consider \textit{inter-sectional group fairness} formulation, where every possible intersection of face component attributes is formulated as one subpopulation, we note that such formulation leads to a subpopulation count that grows exponentially with the number of face component attributes. For instance, for 10 binary biological face attributes and $N$ samples at each intersection of subpopulations, at least $(2^{10}\times N)$ samples are required to evaluate a fairness metric, which is a significant statistical hurdle for data-driven debiasing algorithms because of data scarcity at the intersections. Consequently, we frame our proposed face component fairness as the mean group fairness~\cite{yang_fairness_2020} of face component attributes. Specifically, without loss of generality, we use Equal Opportunity (EO) as our fairness notion for face component level debiasing throughout the paper. 

Formally, based on the definition of Equal Opportunity~\cite{hardt_equality_2016}, we consider a fair classifier to satisfy Equal Opportunity with all chosen face component attributes that minimize true positive rate disparity defined as following:

\begin{definition}[True Positive Rate Disparity (TPRD)]\label{def:TPRD}
    We define the mean True Positive Rate Disparity (TPRD) as:
    \begin{align}
    \begin{split}
    \text{TPRD} = \frac{1}{|\mathcal{A}|}\sum_{A \in \mathcal{A}} \max_{\substack{a_1, a_2 \in \text{Dom}(A), \\ a_1 \neq a_2}} \bigg| P(\hat{Y}=1 \mid A=a_1; Y=1) \\- P(\hat{Y}=1 \mid A=a_2; Y=1) \bigg|,
    \end{split}
    \end{align}
    where \( \mathcal{A} \) is the set of sensitive attributes and \( \text{Dom}(A) \) denotes the set of possible values (realizations) that a sensitive attribute \( A \) can take. TPRD takes value from 0 to 1. The lower the TPRD value, the fairer the model.
\end{definition}

To evaluate TPRD defined above, a straight-forward approach is to evaluate the empirical mean for the conditional probabilities in Definition~\ref{def:TPRD}.
However, face attributes can be inherently correlated due to a number of genetic and perceptual factors~\cite{tanaka_parts_2016}. To verify the correlations between face component attributes, we perform pairwise chi-square independence test on the face attribute annotations of more than 200k celebrity images provided by CelebA dataset~\cite{liu_deep_2015}. For all face component attributes, we observe the majority of attribute pairs manifest a statistical significance of correlations $(p < 0.05)$. For example, we observe statistically significant dependency between \textit{Big Lips} and \textit{Arched Eyebrow}, indicating the need for explicit dependency modeling the calculation of their corresponding TPRD as presented in Section~\ref{dependency_analysis}.

In conclusion, for a face component attribute $A$, the computation of conditional probability $P(\hat{Y}=1|A=a; Y=1)$ in Definition~\ref{def:TPRD} requires estimation of the unknown joint probability distribution $\mathcal{Q} (A, Y, \hat{Y})$ of a face component attribute $A$, prediction target $Y$ and classifier prediction $\hat{Y}$. To model the joint probability distribution $\mathcal{Q} (A, Y, \hat{Y})$, we introduce a Bayesian Network calibrator to achieve more granular and precise fairness evaluation with consideration of the dependencies and prior distributions of face component attributes, which is detailed in Section~\ref{bayesian_calibrator}. During training, we update the Bayesian Network calibrator using Maximum Likelihood Estimation to continuously estimate $\mathcal{Q}(A, Y, \hat{Y})$, providing real-time assessment of prediction bias.

\subsection{Problem Formulation}
To formulate our problem, we start by defining our face component attributes \textit{(e.g., Big Eyes, Big Lips)} to be a set of $K$ discrete random variables $\{A^m\}_{m=1}^K$ for $K$ face attributes. For any downstream face classification task (e.g., smiling detection), a data instance $(x^{(i)}, y^{(i)}, \textbf{a}^{(i)})$ is drawn from dataset $\mathcal{D}$ where $0 \leq i < N$, $x^{(i)} \in \mathcal{X}$ is the feature vector of input face image, $y^{(i)} \in \mathcal{Y}$ is the ground-truth label, and $\textbf{a}^{(i)} \in \mathcal{A}$ is the face component attributes vector of length $K$. Let $\mathcal{D_T}$ be training set, we additionally define $\mathcal{D_F} = \{\mathcal{D}_{val}^{m}\}_{m=1}^K$ to be $K$ micro fairness validation sets with respect to each face component attribute sampled from a larger validation set $\mathcal{D_V}$. Specifically, the sampled micro fairness validation set for attribute $A^m$ denoted as $\mathcal{D}_{val}^{m}$ is a sampled small dataset with equal occurrences of instances with positive and negative values of an attribute $A^m$. Additionally, we denote the positive subset as $Pos (\mathcal{D}_{val}^{m})$ and negative subset as $Neg (\mathcal{D}_{val}^{m})$. The objective is to reduce TPRD that is defined on each attribute in $\mathcal{A}$ and $\mathcal{Y}$ for a chosen downstream task while preserving the task performance of classifier trained on $\mathcal{D_T}$. More formally, our objective can be formulated as a constrained optimization that aims to find the optimal trade-off between downstream face attribute classification accuracy and face component fairness. For a face classifier $f_\theta$ parameterized by $\theta$, the target is to find the best parameter $\theta^*$ in the following optimization: 
\begin{equation}
\theta^* = \arg \min_\theta \mathcal{L}_{task}(\theta, \mathcal{D}) + \lambda \mathcal{L}_{fair}(\theta, \mathcal{D}, \mathcal{A})
\end{equation}
where $L_{fair}$ is a fairness loss constructed from the chosen fairness metric, $L_{task}$ is the task loss and $\lambda$ is some fairness-task performance trade-off parameter under this formulation. 

\subsection{Fairness Optimized Meta-reweighting}

Our methodology implements the "\textit{learning to reweight samples}" objective similar to the overarching principle of "\textit{learning to learn}" in meta-learning by integrating a fairness objective to the sample reweighting process. Specifically, our approach directly incorporates the gradient information from the fairness metrics, thus jointly optimizing the fairness and task performance of the trained classifier. While prior work typically requires a representative fairness validation set~\cite{learning-to-reweight-data, shu_meta-weight-net_2019}, we diverge from them through an individualized assessment for each face component attribute with a micro fairness validation set. The individualized assessment allows for fairness evaluation without having to specify an exemplar set.

In order to integrate fairness directly into the training process, we define two loss functions for the two objectives (\textit{fairness, accuracy}) respectively. We denote $\textbf{v}^{(i)}$ as the $i$-th element of any single column vector $\textbf{v}$. Let $A^m$ be a face component attribute and $\mathcal{D}^m_{val}$ be the corresponding micro fairness validation set, the fairness loss is defined as:
\begin{align}
    \label{eq:lfair}
    \begin{split}
    l_{\text{fair}} (\theta, \mathcal{D}_{\text{val}}^m) = \bigg|E_{(x, \textbf{a}, y=1)\sim \text{Pos}(\mathcal{D}_{\text{val}}^m)} [f_\theta(x)|A^m = \textbf{a}^{(m)}] \\
    - E_{(x, \textbf{a}, y=1)\sim \text{Neg}(\mathcal{D}_{\text{val}}^m)} [f_\theta(x)|A^m = \textbf{a}^{(m)}]\bigg|
    \end{split}
\end{align}
Consequently, we further define the overall independent group fairness on fairness validation sets $\mathcal{D_F}$ based on \textit{Equal Opportunities} principle~\cite{hardt_equality_2016} which is empirically evaluated as TPRD in Definition~\ref{def:TPRD}: 
\begin{equation}
    \mathcal{L}_\text{fair} (\theta, \mathcal{D_F}) = E_{\mathcal{D}^m \sim \mathcal{D_F}} [l_\text{fair} (\theta, \mathcal{D}^m)] 
\end{equation}
For task-specific loss function, we adopt cross-entropy loss widely used for classification tasks, denoted by $\mathcal{L}_{ce}$.

For each batched input, we adopt meta-learning to adaptively learn a weighting vector $w$. At each training time step $t$, we denote weighting vector for current input batch as $w_t$, where we iteratively update weighting vector $w_t$ to $w_{t+1}$ and classifier parameters $\theta_t$ to $\theta_{t+1}$. For our meta-learning process at time step $t$, we start by updating the original classifier $f_{\theta_t}$ to $f_{\theta'}$ using $w_t$ by evaluating the task loss $\mathcal{L}_{ce}$, resulting in an intermediate temporary classifier $f_{\theta'}$. The temporary classifier is then evaluated on $\mathcal{L}_\text{fair}$. As a result, the gradient propagation from $\mathcal{L}_{fair}$ is used to update weight vector $(w_t \rightarrow w_{t+1})$. After this step, the original classifier is updated ($\theta_t \rightarrow \theta_{t+1}$) using the learned weight vector $w_{t+1}$ that can effectively reweigh input data according to our defined fairness loss via gradient descent.

\subsection{Bayesian Network-informed Fairness Calibration for Sample Reweighting}\label{bayesian_calibrator}

With the fairness loss and the optimization algorithm defined above, computing the conditional expectation in Equation~\ref{eq:lfair} typically requires knowledge of joint distribution $\mathcal{Q}(A, Y, \hat{Y})$ of face component attributes, prediction target and classifier prediction. To estimate $\mathcal{Q}(A, Y, \hat{Y})$, we introduce a probability graphical model in our algorithm, specifically a Bayesian Network under the assumption of conditional independence. We denote our Bayesian Network estimated joint probability distribution as $\mathcal{Q^*}(A, Y, \hat{Y})$.

In our approach, to integrate the prior knowledge and the dependencies between face component attributes and the biased predictions, we perform structural learning and parameter estimation on training set $\mathcal{D_T}$. We define our Bayesian Network $\mathcal{B^0}$ parameterized by $\Phi_0$ for a chosen downstream task and a set of face component attributes $\{A^m\}^K_{m=1}$. $\mathcal{B'}_{\Phi_0}$ is trained on data instances $\textbf{a}^{(i)}$ drawn from $\mathcal{D_T}$. For structural learning, we adopt exhaustive search with a K2Score~\cite{koller_probabilistic_2010} criteria that finds the best directed acyclic graph(DAG) structure that fits training data according to K2Score. After structural learning, we adopt graph pruning based on chi-square test of independence and remove any statistically significant independent edge connections. For parameter estimation, we estimate the conditional probability distribution (CPD) in a table representation for each node in the learned structure with a Maximum Likelihood Estimator. 

During initialization, we append a new node of model classifier prediction to Bayesian Network $B_{\Phi_0}$ as a random variable in estimated joint distribution $\mathcal{Q^*}(A, Y, \hat{Y})$ with uniform probability across all possible face component attribute combinations. During training, we update the conditional probability distributions connecting to the prediction node while preserving prior information of attributes with a maximum likelihood estimator for every $N$ steps of training where $N$ is a hyper-parameter controlling the frequency of belief update.

\subsubsection{Fairness Loss Evaluation}

In this subsection, we describe in detail how we evaluate the fairness loss function $\mathcal{L}_\text{fair}(\theta, \mathcal{D_F})$ for a given model $f_\theta$ and fairness validation sets $\mathcal{D_F}$. From Equation~\ref{eq:lfair}, we observe that $l_\text{fair}(\theta, \mathcal{D}_\text{val}^m)$ is the empirical evaluation of true positive rate disparity. Denoting the TPRD for attribute $A^m$ as $\delta^m$, the evaluation of $\delta^m$ requires the conditional probability $P(\hat{Y}=1|A=a)$ given an attribute observation $A=a$. According to Bayes Rule, we can expand this conditional probability as follows:
\begin{align}
\begin{split}
P(\hat{Y}=1| A=a) &= \frac{P(A = a|\hat{Y} = 1)P(\hat{Y} = 1)}{P(A = a)} \\
  &= P(\hat{Y} = 1) \mathbf{Z}
\end{split}
\end{align}
where 
\[
\mathbf{Z} = \frac{P(A = a|\hat{Y} = 1)}{P(A = a)}
\]
In our approach, we empirically evaluate $P(\hat{Y} = 1)$ from the classifier confidence of a sample belonging to the positive class, while $\mathbf{Z}$ is a quantity that we refer to as Bayesian calibrator. By definition, the implication of the Bayesian calibrator $\mathbf{Z}$ can be understood as a likelihood ratio or as a measure of how positive prediction changes our belief about the attribute $A = a$. Specifically, we evaluate $\mathbf{Z}$ using the trained Bayesian Network $\mathcal{B}_\phi$ that encodes dependency and prior beliefs of face component attributes. Specifically, to evaluate a probability evaluation $P(A = a)$, we adopt variable elimination to query the trained Bayesian Network to obtain $P(A=a)$. To obtain $P(A=a|\hat{Y}=1)$, we additionally set positive prediction observation $\hat{Y}=1$ in the trained Bayesian Network before querying $\mathcal{B}_\phi$. 
\begin{table*}[t]
  \caption{Face Component Fairness on CelebA Dataset (\%) for 3-Attribute Setting (\textit{big lips, arched eyebrows, big nose}) and 5-Attribute Setting (\textit{big lips, arched eyebrows, big nose, double chin, no beard})}
  \label{tab:main}
  \centering
  \renewcommand{\arraystretch}{1.2} 
  \setlength{\tabcolsep}{6pt} 
  \begin{tabular}{cc|ccc|ccc}
    \hline
    \hline
    \multirow{2}{*}{\textbf{Task}} & \multirow{2}{*}{\textbf{Methods}} & \multicolumn{3}{c|}{\textbf{3 Attributes}} & \multicolumn{3}{c}{\textbf{5 Attributes}} \\
    \cline{3-5} \cline{6-8}
    & & \textbf{Accuracy} $\uparrow$ & \textbf{DIG} $\downarrow$ & \textbf{TPRD} $\downarrow$ & \textbf{Accuracy} $\uparrow$ & \textbf{DIG} $\downarrow$ & \textbf{TPRD} $\downarrow$ \\
    \hline
    \multirow{7}{*}{\textbf{Attractiveness}}
    & Vanilla & 80.13 & 13.71 & 10.79 & 80.13 & 23.24 & 16.84 \\
    & Random & 79.90 & 12.30 & 9.76 & 79.90 & 20.17 & 13.56 \\
    & FORML & 80.06 & 12.47 & 10.08 & 80.04 & 15.50 & 12.66 \\
    & KD & 79.96 & 12.31 & 9.70 & 79.96 & 25.80 & 19.16 \\
    & Adversarial & 79.82 & 20.14 & 14.74 & 79.35 & 27.21 & 18.89 \\
    & InfoFair & 79.31 & 14.41 & 11.22 & 79.06 & 20.54 & 17.46 \\
    & BNMR (Ours) & 79.75 & \textbf{10.38} & \textbf{8.67} & 80.13 & \textbf{13.43} & \textbf{10.76} \\
    \hline
    \multirow{7}{*}{\textbf{Smiling}}
    & Vanilla & 92.29 & 5.66 & 5.13 & 92.29 & 5.12 & 4.60 \\
    & Random & 92.47 & 5.94 & 5.35 & 92.47 & 5.13 & 4.60 \\
    & FORML & 92.32 & 5.23 & 4.78 & 92.23 & 4.71 & 4.17 \\
    & KD & 91.97 & 5.94 & 5.32 & 91.97 & 4.36 & 5.94 \\
    & Adversarial & 91.24 & 5.45 & 4.91 & 91.59 & 4.16 & \textbf{3.80} \\
    & InfoFair & 91.33 & 5.24 & 4.68 & 91.34 & 4.81 & 4.25 \\
    & BNMR (Ours) & 92.00 & \textbf{4.94} & \textbf{4.12} & 92.10 & \textbf{2.50} & 3.88 \\
    \hline
    \hline
  \end{tabular}
\end{table*}

\subsubsection{Normalized Sample Re-weighting}
To increase the stability and flexibility of our approach, we adopt softmax normalization on the weight vector before applying sample reweighting. Specifically, our normalization scheme for a weight vector $w$ is defined as an element-wise function: 
\begin{align}
\label{eq:softmax}
\begin{split}
\rho(w^{(i)}, \tau) = \frac{exp(\frac{w^{(i)}}{\tau})}{\sum_j exp(\frac{w^{(j)}}{\tau})}
\end{split}
\end{align}
where we introduce a temperature parameter $\tau$ that controls the sharpness of the normalized weight vector.


\section{Experiments}
In this section, we present experiments investigating face component fairness, the effectiveness of BNMR, and the relationship between face component fairness and demographic fairness. To further assess the performance and efficacy of our method, we also conduct an ablation study and a weight sensitivity analysis for sample reweighting.

\subsection{Face Component Fairness Analysis}\label{dependency_analysis}
In this section, we analyze bias at the face component level and the interdependence among facial attributes using the CelebA dataset~\cite{liu_deep_2015}. We first observe significant class imbalance across many face component attributes, which could affect model performance and fairness. For instance, despite the importance of eyebrows in facial recognition tasks~\cite{Lestriandoko_contribution_2022}, only $26.7\%$ of samples exhibit \textit{arched eyebrows}, suggesting an under-representation of this feature (examples are demonstrated in Figure~\ref{fig:examples}).

Furthermore, we observe complex inter-dependencies among different face component attributes by examining network relationship graphs generated from a Bayesian Network extracted from CelebA dataset. As shown in Figure~\ref{fig:dependency}, when considering face component attributes (\textit{big lips, arched eyebrows, big nose, double chin, no beard}), the conditional probability distributions extracted from data indicate complex inter-dependencies such as $(\phi(0.2434): \textit{Big Lips} \rightarrow \textit{Arched Eyebrow}), (\phi(0.2538): \textit{Big Nose} \rightarrow \textit{No Beard})$ and $(\phi(0.2984):\textit{Big Nose} \rightarrow \textit{Double Chin})$, where $\phi(\cdot)$ indicates $\phi$ value obtained from chi-square test of independence. Such inter-dependencies could potentially introduce conflicts in fairness optimization for different sensitive attributes. For example, there is a negative correlation between \textit{No Beard} and \textit{Big Nose} $(\phi=0.2538)$. When debiasing with respect to \textit{No Beard} using sample reweighting, larger sample weights assigned to instances with \textit{No Beard} would potentially result in uncontrolled increasing of sample weights for instances without \textit{Big Nose}, which could lead to a more significant bias against the \textit{Big Nose}. Therefore, we conclude that face component attributes are highly correlated, and do not meet the assumption that sensitive attributes are independent~\cite{pmlr-v80-kearns18a}. In our approach, we use a Bayesian Network calibrator to explicitly model such inter-dependencies between different face components, facilitating a more accurate fairness loss evaluation and a more effective bias mitigation.

\subsection{Experiment Setting}
We evaluate our approach on a large scale human face dataset CelebA~\cite{liu_deep_2015}, which is currently the only publicly available dataset with face component-level annotations. CelebA dataset contains 202,599 human faces with face attribute annotations. In order to improve generalizability of our approach within the constraints of existing data, we randomly sample 50000 images of the training data to simulate different biased training inputs following prior work~\cite{zeng_fairness-aware_2023}.
Among the annotated attributes, we choose binary attributes smile and attractiveness separately as our prediction target following Zeng et al.~\cite{zeng_fairness-aware_2023} where $y=1$ indicates "smiling" or "attractive" and $y=0$ otherwise. We note that "attractiveness" classifiers are more likely to suffer from bias due to the subjectivity of target label~\cite{shen_fooling_2017}, whereas smiling detection is a simpler task since it is more distinguishable based on face features~\cite{zeng_fairness-aware_2023}. For face component attributes, we choose 5 attributes that are highly visible and distinct, which may have a larger impact on decision-making in face attribute classification~\cite{Lestriandoko_contribution_2022}. Specifically, we selected the following binary facial component attributes: \textit{big lips, arched eyebrows, big nose, double chin, no beard} To investigate how the number of chosen facial component attributes affects the debiasing performance, we conducted experiments using either three attributes (\textit{big lips, arched eyebrows, big nose}) or all five attributes. 

It is worth noting that in real-world deployment, the selection of face component attributes and their parameters should be informed by fairness auditing to ensure they accurately represent existing biases and mitigate label bias (e.g., assessing whether ``big lips'' contributes to biased classification). While an exhaustive selection process could provide additional insights, it lies beyond the scope of this study, as our focus is on demonstrating the BNMR's adaptability and performance under controlled attribute configurations.

Throughout our experiment, we adopt the train, test, and validation split provided by CelebA dataset. All input images are normalized using mean and standard deviation from ImageNet dataset~\cite{deng_imagenet_2009}. For our experiments, we use \textit{lightCNN}~\cite{wu_light_2018} as our backbone for face attribute classification given its wide adoption in face attribute classification~\cite{zeng_fairness-aware_2023,zeng_on_2023}. In each training, we choose the most accurate model checkpoint on the validation set for fairness evaluation. For hyper-parameter selection of each experiment, we conducted independent grid searches (Learning Rate: [1e-5,1e-4,1e-3]; Prior sample size for Bayesian Network updates: [40,80,160]; $\tau$: [0.1, 0.2, ..., 1.0]), guided by hyper-parameters reported in prior work~\cite{wu_light_2018}. Our code is made publicly available~\footnote{https://github.com/yliuaa/BNMR-FairCompFace.git} with the best hyper-parameters reported in Table~\ref{tab:hyperparameters}.

\begin{table}[h]
\centering
\begin{tabular}{ll}
\hline
\textbf{Parameter}         & \textbf{Value} \\ \hline
Batch size                 & 16             \\
Fairness validation size   & 20             \\
Learning rate              & $1 \times 10^{-4}$ \\
$\tau$                     & 0.9            \\
Prior sample number for Bayesian Network updates & 80 \\ \hline
\end{tabular}
\caption{Hyperparameter settings used in the experiments.}
\label{tab:hyperparameters}
\end{table}

For evaluation metrics, we consider Disparate Impact Gap (DIG) and True Positive Rate Disparity (TPRD) from Definition~\ref{def:TPRD} to be our fairness metrics with DIG comparing the ratios of positive outcomes between them and TPRD examining the difference in true positive rates across groups.

\begin{definition}[Disparate Impact Gap (DIG)]
    We define the mean Disparate Impact Gap (DIG) as:
    $$\text{DIG} = \frac{1}{|\mathcal{A}|}\sum_{A \in \mathcal{A}} \max_{a_1, a_2 \in \text{Dom}(A); \, a_1 \neq a_2} \bigg| 1 - \frac{P(\hat{Y}=1 \mid A=a_1, Y=1)}{P(\hat{Y}=1 \mid A=a_2, Y=1)} \bigg|,$$
    where \( \mathcal{A} \) is the set of sensitive attributes and \( \text{Dom}(A) \) denotes the set of possible values (realizations) that a sensitive attribute \( A \) can take. DIG takes values from 0 to 1, with lower DIG values indicating a fairer model.
\end{definition}

\subsection{Baselines}
In our experiments, we select the best performing baselines from recent literature that have open-source code or reproducible results. Specifically, we evaluate the effectiveness of our approach by comparing it to two fixed naive baselines: vanilla and random re-weighting training and two closely related state-of-the-art debiasing algorithms: FORML~\cite{learning-to-reweight-data} and Fairness without demographics through knowledge distillation (KD)~\cite{chai2022fairness}. Additionally, we consider state-of-the-art multi-attribute debiasing competitors InfoFair~\cite{kang_infofair_2022} and adversarial debiasing scheme~\cite{zhang2018mitigating}. For all baselines, we adopt grid-search to find their best hyper-parameters on the validation set provided by CelebA. 

\subsection{Main Results}
Our main results are summarized in Table~\ref{tab:main}. In our experiments on attractiveness classification and smiling detection, our proposed BNMR approach consistently achieves superior fairness metrics while maintaining competitive accuracy compared to existing methods, indicating its effectiveness in enhancing fairness across different tasks without compromising accuracy.

In attractive classification, BNMR achieves the lowest mean DIG and mean TPRD in both 3 and 5 attribute settings, indicating a substantial reduction in disparity where the reduction in mean DIG for 5-attribute attractiveness classification is particularly notable ($9.81\%$). Moreover, for both 3 and 5 attributes, BNMR maintains or matches the highest accuracy, showing such fairness improvements do not come at the cost of significant accuracy lost.

For smiling detection with 3 attributes, BNMR reduced mean DIG by 1.92\%, outperforming methods like Adversarial debiasing and InfoFair with only a marginal decrease in accuracy (less than $0.5\%$) compared to the most accurate baseline (\textit{Vanilla}). The higher accuracy of the vanilla classifier without bias mitigation suggests that all bias mitigation approaches involve an accuracy-fairness trade-off. For smiling detection with 5 attributes, BNMR achieves the lowest mean DIG and second lowest mean TPRD while maintaining competitive accuracy. We notice that although adversarial debiasing achieves slightly lower TPRD with 5 attributes, BNMR significantly outperforms it in mean DIG, indicating a better overall fairness performance.

In addition to the fairness improvements, we observe that BNMR achieves a more significant fairness improvement with a larger number of face component attributes,
showing the scalability of BNMR's bias mitigation. Specifically, compared to the second-best baselines, BNMR achieves a clear DIG improvement $(1.92\% \rightarrow 2.07\%, 0.30\% \rightarrow 1.66\%)$ when increasing from 3 to 5 attributes for attractiveness and smiling detection tasks. Furthermore, we notice that BNMR achieves better accuracy—improving $(79.75\% \rightarrow 80.13\%,\ 92.00\% \rightarrow 92.10\%)$—when increasing from 3 to 5 attributes for the two face attribution classification tasks. The possible reason is a larger number of facial component attributes often leads to a more accurate conditional probability modeling generated by Bayesian Network calibrator, which subsequently provides better fairness guidance.

\begin{table*}[t]
  \caption{Ablation Study for Attractiveness and Smiling (\%)}
  \label{tab:ablation}
  \centering
  \renewcommand{\arraystretch}{1.2} 
  \setlength{\tabcolsep}{4pt} 
  \begin{tabular}{c|cccc|cccc}
    \hline
    \hline
    \multirow{2}{*}{\textbf{Method}} & \multicolumn{3}{c}{\textbf{Attractiveness}} && \multicolumn{3}{c}{\textbf{Smiling}} \\
    \cline{2-4} \cline{6-8}
     & \textbf{Acc.} $\uparrow$ & \textbf{DIG}$\downarrow$ & \textbf{TPRD} $\downarrow$ &&
    \textbf{Acc.} $\uparrow$ & \textbf{DIG}$\downarrow$ & \textbf{TPRD} $\downarrow$ \\
    \midrule
    w/o Softmax Normalization & 80.35 & 14.86 & 12.30 && 91.90 & 3.89 & 4.26 \\
    w/o Online Update & 80.39 & 15.41 & 19.86 && 91.83 & 4.82 & 4.35 \\
    w/o Bayesian Calibration & 79.96 & 19.16 & 25.80 && 92.23 & 4.17 & 4.71 \\
    w/o Reweighting & 80.13 & 16.84 & 23.24 && 92.29 & 4.60 & 5.12 \\
    Ours & 80.13 & \textbf{10.76} & \textbf{13.43} && 92.10 & \textbf{3.88} & \textbf{2.50} \\
    \hline
    \hline
  \end{tabular}
  \begin{center}
    \textbf{Abbreviations:} w/o = without, DIG = Demographic Intersectional Gap, TPRD = True Positive Rate Difference.
  \end{center}
\end{table*}


\subsection{Ablation Study}
As shown in Table~\ref{tab:ablation}, we conduct an ablation study under the more complex 5 face component attributes setting by removing different components in our pipeline, specifically online updates of Bayesian Network, tempered softmax normalization of weight vector, Bayesian Network guidance, and sample re-weighting. The results across two tasks suggest that each of our training designs contributes to a more desirable accuracy-fairness trade-off, with Bayesian calibration plays the most crucial role in bias mitigation.

In our ablation study, we observe that \textit{w/o softmax normalization} exhibits more bias in both tasks. Such a performance drop can be attributed to the uncontrolled scale of weight vector guided completely by back-propagation. It is worth noting that the optimal strength of our bias mitigation can be obtained through simple parameter tuning of tempered softmax normalization.
When removing Bayesian calibration, the bias mitigation no longer has information for face component attributes' inter-dependencies and belief of current model bias. Consequently, we observe deteriorated fairness performance. When removing online update of Bayesian Network calibrator, belief of model bias is not updated although the inter-dependencies are encoded. The outdated beliefs about model bias lead to misaligned bias evaluations in the fairness loss, which hinders effective bias mitigation. Finally, we observe that without sample reweighting, the results from vanilla training exhibit the highest level of bias compared to other ablated baselines. The deteriorated fairness of \textit{w/o sample reweighting} baseline shows the overall effectiveness of our reweighting-based debiasing scheme.

\subsection{Face Component Fairness and Demographic Fairness}
Following the convention of prior work~\cite{wang2022fairness, zhang2022fairness, kim2019multiaccuracy, Quadrianto_2019_CVPR, Wang_2020_CVPR}, we use \textit{male} as our demographic attribute and investigate the correlations among three groups of variables of interest: face component attributes (\textit{big lips, arched eyebrows, big nose, double chin, no beard}), demographic attributes (\textit{male}) and prediction target (\textit{smiling, attractiveness}) through comprehensive statistical testings. We first carry out pairwise chi-square test of independence among the variable groups to check the existence of significant correlations. We report our $\phi$-coefficient given by $\phi = \sqrt{\frac{\mathcal{X}^2}{n}}$ in Table~\ref{tab:chi-square} where $\phi$ takes range from 0 to 1 with higher value indicating stronger correlation. In addition, we report demographic fairness of our baselines under 5 face component attribute setting in Table~\ref{tab:demographic}.

In Table~\ref{tab:chi-square}, all reported independence tests show statistical significance with $p \ll 0.05$. Regarding the strength of correlations, our analysis reveals that the selected face component attributes exhibit moderate to strong correlations with \textit{male}, indicated by $\phi$ values in $[0.1675, 0.5222]$. Additionally, the correlations between face component attributes and the two prediction targets are weak to moderate, indicated by $\phi$ values in $[0.0129, 0.1384]$ and $[0.1976, 0.2771]$ respectively. Among the five chosen face component attributes, each attribute shows a stronger correlation with \textit{male} than with the prediction targets (i.e., smiling and attractiveness). Furthermore, we observe that \textit{male} exhibits a stronger correlation to the prediction targets than all face component attributes.

We explain the insights of the above observations through the mutual information in information theory. Specifically, for a classifier $f^*(X)$ to be fair with respect to a sensitive attribute (demographic or face component) $A$ and accurate with respect to prediction target $Y$, the mutual information $I(f^*(X), A)$ needs to be small and $I(f^*(X), Y)$ to be large. Under the assumption that $f^*(X)$ is an optimal predictor of $Y$ and thus captures all relevant information from $X$ for accurate prediction, it follows that $f^*(X)$ encodes information about $Y$. As shown in Table~\ref{tab:chi-square}, the stronger correlations between \textit{male} and prediction targets compared to face component attributes indicate an inherent higher value of $I(A_{male}, Y)$. Therefore, to debias such a predictor, it will need to ignore/remove more information \textit{male} compared to face component attributes that are encoded in $Y$. This explains why the gender fairness is a more difficult optimization target than the face component attributes.

The strong correlation between demographic attribute (\textit{gender}) and face component attributes suggest that the removal of mutual information between face component attributes and prediction target can potentially benefit the removal of mutual information between demographic attribute and prediction target, thus improving demographic fairness. We report demographic fairness of our experiments in Table~\ref{tab:demographic}. \textbf{We observe that BNMR, while being the fairest model on face component attributes, is also the fairest model on demographic attribute.} We conclude that, for our selected face component attributes, face component fairness can serve as a reasonable proxy for demographic fairness.

\begin{table}[t]
  \caption{$\phi$ Values from Chi-square Test of Independence of Variables of Interest}
  \label{tab:chi-square}
  \centering
  \renewcommand{\arraystretch}{1.2}
  \begin{tabular}{c|ccc}
    \hline
    \hline
    \textbf{Attributes} & \textbf{Male} & \textbf{Smiling} & \textbf{Attractiveness} \\
    \midrule
    \textit{Arched Eyebrow} & 0.4080 & 0.0938 & 0.2506 \\
    \textit{Big Lips} & 0.1675 & 0.0129 & 0.0625 \\
    \textit{Big Nose} & 0.3693 & 0.1009 & 0.2771 \\
    \textit{Double Chin} & 0.2075 & 0.1001 & 0.2090 \\
    \textit{No Beard} & 0.5222 & 0.1128 & 0.1976 \\
    \midrule
    \textbf{Male} & 1.0000 & \textbf{0.1384} & \textbf{0.3944} \\
    \hline
    \hline
  \end{tabular}
\end{table}

\begin{table}[t]
  \caption{Demographic Fairness (\textit{Gender}) of Both Prediction Targets (\textit{Attractiveness and Smiling}) with 5 Face Component Attributes (\%)}
  \label{tab:demographic}
  \centering
  \renewcommand{\arraystretch}{1.2}
  \begin{tabular}{c|cc|cc}
    \hline
    \hline
    \multirow{2}{*}{\textbf{Methods}} & \multicolumn{2}{c}{\textbf{Attractiveness}} & \multicolumn{2}{c}{\textbf{Smiling}} \\
    \cline{2-3} \cline{4-5}
     & \textbf{DIG} $\downarrow$ & \textbf{TPRD}$\downarrow$ & \textbf{DIG}$\downarrow$ & \textbf{TPRD}$\downarrow$ \\
    \hline
    Vanilla & 34.34 & 29.90 & 3.69 & 3.62 \\
    Random & 34.81 & 30.74 & 4.13 & 3.82 \\
    FORML & 30.49 & 26.78 & 4.85 & 4.36 \\
    KD & 32.09 & 28.24 & 3.97 & 3.67 \\
    Adversarial & 36.18 & 32.20 & 2.90 & 2.67 \\
    InfoFair & 32.55 & 29.20 & 2.95 & 2.70 \\
    \hline
    \textbf{BNMR (Ours)} & \textbf{23.87} & \textbf{21.97} & \textbf{1.99} & \textbf{1.67} \\
    \hline
    \hline
  \end{tabular}
\end{table}


\subsection{Weighting Sensitivity Analysis}

\begin{figure}[htbp]
    \centering
    \includegraphics[width=\linewidth]{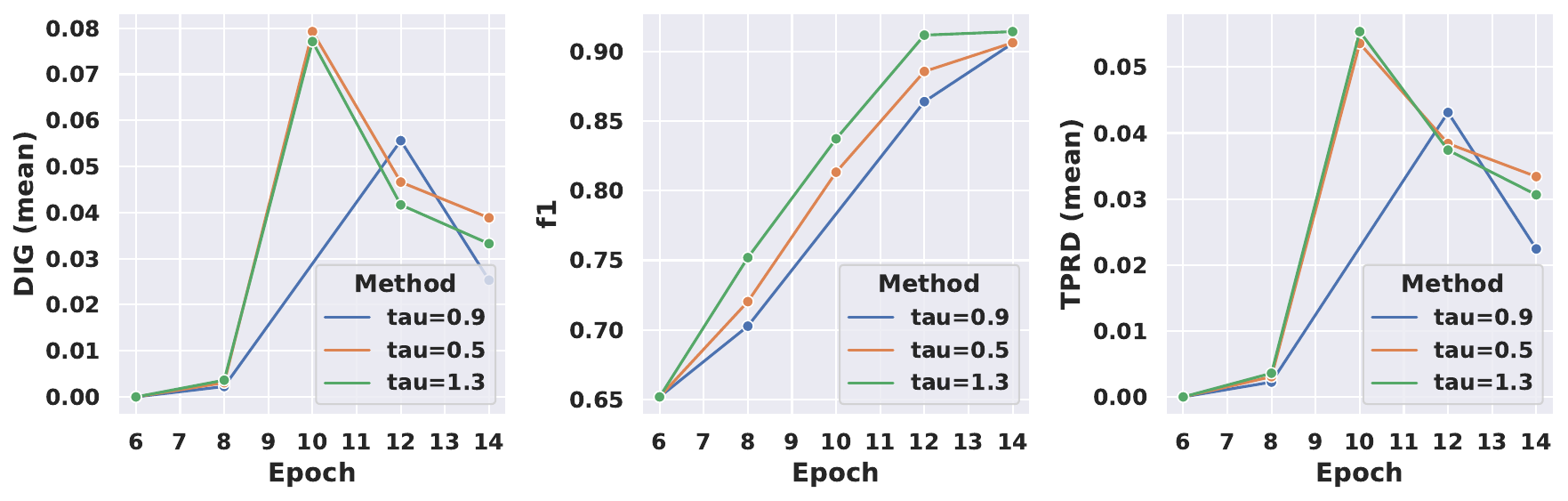}
    \caption{Weighting sensitivity analysis on $\tau$ in re-weighting. The value $\tau=0.9$ provides the best fairness result. A value of $\tau$ that is either excessively high or low diminishes the effectiveness of bias mitigation efforts.}
    \label{fig:parameter}
\end{figure}

We evaluate the effect of parameter $\tau$ that adjusts the sharpness of re-weighting vector by taking the learning curves of 5 attributes of smiling detection as an example. The results are shown in Figure~\ref{fig:parameter}. Empirically in our experiments, we observe that the optimal $\tau$ values are found between 0 and 2. As defined in Equation~\ref{eq:softmax}, we notice $\tau>1$ would reduce the differences between the sample weights, leading to a more equal gradient update for each sample. As a result, the training is more similar to vanilla training. In contrast, $\tau<1$ would enhance the difference between sample weights, indicating a more aggressive sample reweighting.

At $\tau=0.9$, we observe best fairness in terms of both mean DIG and TPRD, suggesting a slightly sharper weight distribution than default ($\tau = 1$) can aid effective fairness re-weighting by giving more weights to biased samples while not overly emphasizing any potential bias present in the data. With an appropriate selection of $\tau$, we observe that although the training process is slowed down, BNMR does not pose significant differences on the final convergence point (\textit{Epoch=14} in Figure~\ref{fig:parameter}). Such a unified convergence point indicates that BNMR does not require significantly additional training time to debias the training process.

\subsection{L1- vs. L2-norm TPRD Objective}
We note that, unlike L2-norm loss which is strongly convex, L1-norm disparity loss used in Equation~\ref{eq:lfair} does not inherently penalize disparity redistribution across attributes. However, we observe L1-norm loss yield better accuracy-fairness trade-off in our experiments, as illustrated in Table~\ref{tab:l1andl2}. In particular, in our fairness loss evaluation, a Bayesian network calibrates probability estimates based on learned dependencies, ensuring that probability adjustments are not made independently for each attribute. Specifically, in a naive setting without Bayesian calibration, reducing the disparity of one attribute ($A_1$) by increasing another ($A_2$) is unconstrained, leading to free disparity trade-offs. With Bayesian calibration, this redistribution is restricted by the learned dependency structure, meaning the increase in $A_2$ disparity is bounded. As a result, the total fairness loss ensures fairness adjustments remain probabilistically justified rather than purely numerical.

\begin{table}[h]
\centering
\caption{Comparison of L1-loss and L2-loss for fairness optimization. Results are shown for two prediction targets (Smiling and Attractiveness) across different fairness metrics.}
\begin{tabular}{lcccc}
\hline
\hline
\multicolumn{5}{c}{\textbf{Smiling}} \\
\cmidrule(lr){2-5}
\textbf{Method} & 3-DIG $\downarrow$ & 3-TPRD $\downarrow$ & 5-DIG $\downarrow$ & 5-TPRD $\downarrow$ \\
\midrule
L2-loss & 5.29 & 4.75 & 4.43 & 4.94 \\
L1-loss & 4.94 & 4.12 & 2.50 & 3.88 \\
\addlinespace[1ex]
\hline
\multicolumn{5}{c}{\textbf{Attractiveness}} \\
\cmidrule(lr){2-5}
\textbf{Method} & 3-DIG $\downarrow$ & 3-TPRD $\downarrow$ & 5-DIG $\downarrow$ & 5-TPRD $\downarrow$ \\
\midrule
L2-loss & 12.38 & 14.13 & 15.05 & 15.44 \\
L1-loss & 10.38 & 8.67 & 13.43 & 10.76 \\
\hline
\hline
\end{tabular}
\label{tab:l1andl2}
\end{table}

\section{Discussion}
\textbf{Choosing attractiveness as a prediction target.} It is worth noting that attractiveness label is highly subjective and could encode inherent labeling bias. Imposing a fairness criterion on a prediction task that has biased labels does not address the underlying bias in those labels. However, our reason for attractive prediction, was that attractiveness labels in popular face datasets are notoriously prone to imbalance and subjectivity~\cite{fooling_shen}, making them a good stress-test for fairness interventions, as seen in prior work~\cite{KOLLING2023173,fooling_shen,zeng_fairness-aware_2023,zeng_adversarial}. Our focus was solely on evaluating how well our method could correct bias. Furthermore, our choice of dataset was dictated by our focus on face component fairness. To our best knowledge, CelebA is the only publicly available dataset with face component-level annotations~\cite{celeba}, and our task selection was constrained by available labels. We have examined all non-face components and non-demographic labels in CelebA and found that only smiling and attractiveness exhibited significant vanilla training bias suitable for our evaluation.

Indeed, attractiveness is highly subjective and culturally constructed, and human judgments of attractiveness are known to reflect social biases~\cite{lookism, celeba_lingenfelter}. We fully recognize the limitation here: if the labels are themselves biased or tainted by prejudice, mitigating bias in the model’s output cannot fully solve the deeper issue. This is a general challenge in fairness research: bias mitigation algorithms typically assume an unbiased ground truth and focus on model parity~\cite{jiang2019identifyingcorrectinglabelbias}. In future work, we will explore applications with more objectively defined ground truth with our own face component annotations.

\textbf{Clarifying the Significance of Face Component Fairness.}
The motivation and significance of our face component fairness setting are twofold: (1) Facial component attributes (e.g., lip fullness, nose shape, eye size) independently contribute to biases of face recognition models even within the same demographic group, necessitating explicit bias mitigation beyond demographic attributes; (2) Facial component fairness are connected to demographic parity in its definition and bias mitigation optimization.

Evaluating fairness based on facial component attributes is directly tied to documented real-world biases linked to morphological traits (e.g., eye shape, lip fullness), which have caused algorithmic discrimination independently of demographic groups~\cite{afrocentric_features,eberhardt_looking,Phenotypicality}. Such component-level bias is also observed in face recognition systems~\cite{time2025facedetection}. Furthermore, there is a growing recognition that fairness should account for face component attributes to avoid overlooking nuanced biases~\cite{lookism,qiu2024gone,yucer1,yucer2}.

Face component fairness is connected to demographic parity in its definition and as an optimization goal. Prior work argues that broad demographic labels are too coarse and suggests evaluating bias at the level of specific face components (e.g., eye shape, nose shape) for a more explainable and actionable view of demographic bias~\cite{qiu2024gone}. From an optimization perspective, face components have also been utilized in bias mitigation while preserving user privacy and have demonstrated effectiveness in reducing bias~\cite{non_sensitive_esteban,yucer3}.

While our study centers on face component fairness, the underlying methodology is broadly applicable to other domains with fine-grained multi-attribute biases. Given the significance of face component fairness and the lack of prior discussion in this area, our work represents a unique and essential step toward addressing real-world biases associated with morphological traits.

\textbf{Real-world context of our work.}
Facial feature related bias is not merely a technical issue but a broader real-world societal concern requiring interdisciplinary perspectives. For instance, Eberhardt et al. provide strong evidence that facial features alone influence sentencing decisions, with individuals possessing certain features (e.g., thick lips, broad noses) receiving disproportionately higher death sentences~\cite{eberhardt_looking,seeingblack}. The widespread adoption of AI-driven facial recognition systems across various domains (e.g., recidivism prediction and justice systems~\cite{zeng_adversarial}) risks perpetuating such biases~\cite{broussard2023more}.

Specifically, real-world face recognition systems demonstrate significant harm caused by facial component-level biases. In particular, biased face recognition systems have led to wrongful arrests, unfair sentencing, and racial profiling~\cite{survey,rice1996race}. In hiring, AI-driven applicant screening tools may systematically disadvantage individuals based on facial attributes, further entrenching social inequalities~\cite{recruit_practices}. Without intervention, these biases may be embedded in automated decision-making systems, exacerbating existing societal disparities. 

\textbf{Limitations of our approach.}
While our method effectively improves face component fairness, we observe that sampling fair validation sets
introduces minor additional computational overhead. However, our approach of sample-reweighting does not introduce
extra parameters to the model, making it remain as efficient as vanilla training during inference. Another limitation of
our method is the reliance on the quality of the initial Bayesian Network learned, which could reduce the effectiveness
of our scheme in scenarios where access to extensive labeled data for face component attributes is limited. Addressing
this challenge could involve exploring semi-supervised learning techniques or leveraging synthetic data to enhance the
Bayesian Network’s robustness and applicability. In addition, future work could explore effective methods for selecting
face component attributes as proxies for demographic attributes in bias mitigation, aiming to maximize user privacy
while ensuring fairness. Moreover, the development of additional evaluation benchmarks for face component fairness
could further advance this research area, enabling more comprehensive and well-rounded assessments of BNMR.

\section{Conclusion}
In this paper, we introduce face component fairness, a novel fairness notion defined on biological face attributes. We also develop a meta-learning based approach BNMR to explore the trade-off between face component fairness and task performance. We evaluate our approach on a large scale and publicly available human face dataset and compare our scheme to several closely related debiasing baselines. Experiment results suggest that our method can achieve fair classification on all chosen face component attributes for a smiling detection task. We also identify a positive correlation between face component fairness and gender fairness. 

\begin{acks}
This research is supported in part by the National Science Foundation under Grant No. CNS-2427070,  IIS-2331069,  IIS-2202481, IIS-2130263, CNS-2131622. The views and conclusions contained in this document are those of the authors and should not be interpreted as representing the official policies, either expressed or implied, of the U.S. Government. The U.S. Government is authorized to reproduce and distribute reprints for Government purposes notwithstanding any copyright notation here on.
\end{acks}

\bibliographystyle{ACM-Reference-Format}
\bibliography{ref}

\newpage
\appendix
\section{Case Study: Face Component Bias Analysis}
\begin{figure}[t]
    \centering
    \begin{subfigure}[b]{0.15\textwidth} 
        \centering
        \includegraphics[width=\textwidth]{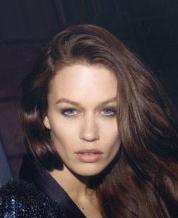}
        \label{fig:1}
    \end{subfigure}
    \hspace{-0.5em} 
    \begin{subfigure}[b]{0.15\textwidth}
        \centering
        \includegraphics[width=\textwidth]{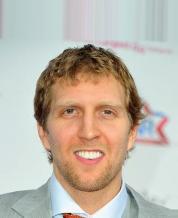}
        \label{fig:2}
    \end{subfigure}
    \hspace{-0.5em} 
    \begin{subfigure}[b]{0.15\textwidth}
        \centering
        \includegraphics[width=\textwidth]{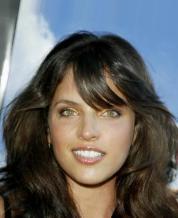}
        \label{fig:3}
    \end{subfigure}

    \begin{subfigure}[b]{0.15\textwidth}
        \centering
        \includegraphics[width=\textwidth]{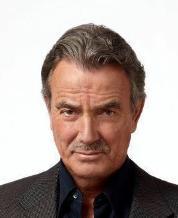}
        \label{fig:4}
    \end{subfigure}
    \hspace{-0.5em} 
    \begin{subfigure}[b]{0.15\textwidth}
        \centering
        \includegraphics[width=\textwidth]{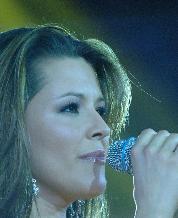}
        \label{fig:5}
    \end{subfigure}
    \hspace{-0.5em} 
    \begin{subfigure}[b]{0.15\textwidth}
        \centering
        \includegraphics[width=\textwidth]{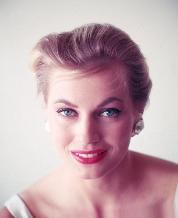}
        \label{fig:6}
    \end{subfigure}
    
    \caption{Examples of false smiling detections caused by the spurious correlation with ``arched eyebrows'' across different genders. These cases highlight the presence of face component bias, where the feature ``arched eyebrows'' disproportionately contributes to incorrect predictions, illustrating the need to address such biases in smiling detection models.}
    \label{fig:examples}
\end{figure}

To motivate our work, we provide a set of case studies showcasing the false classifications with a specific face component attribute (e.g., \textit{arched eyebrows}) across different demographic attributes (e.g., \textit{gender}). As illustrated in Figure~\ref{fig:examples}, we focus on the task of smiling detection to demonstrate how the presence of a facial component attribute, such as \textit{arched eyebrows}, can lead to biased outcomes. We observe that, in the smiling detection task, models often establish a spurious correlation between \textit{arched eyebrows} and smiling, resulting in systematic misclassifications.

Such biases are not distributed uniformly across demographic groups. For instance, females tend to experience higher rates of false positives, suggesting that the interplay between the facial component (\textit{arched eyebrows}) and the demographic attribute (\textit{gender}) exacerbates the model’s errors. This observation highlights the compounding nature of facial component and demographic biases, which together lead to unfair outcomes in smiling detection. Consequently, we highlight the importance of face component fairness. Addressing face component-level biases is critical for developing fairness-aware models that ensure equitable performance.

\section{Learned Bayesian Networks}

\begin{figure}[h]
    \centering
    \includegraphics[width=\columnwidth]{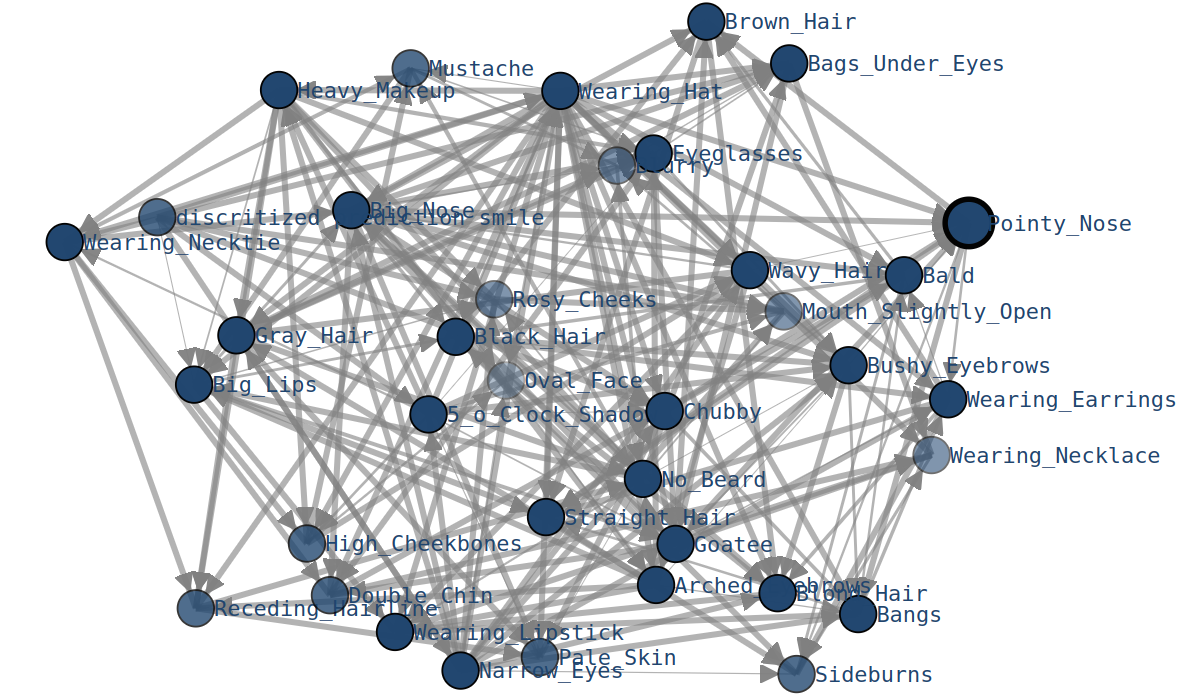}
    \caption{Bayesian Network structure learned for all face attributes on all images in CelebA dataset. The dense connections represent intricate interdependencies, emphasizing the need for probabilistic reasoning when debiasing with respect to face component attributes.}
    \label{fig:bayesian_network}
\end{figure}

\begin{figure}[h]
    \centering
    \includegraphics[width=\columnwidth]{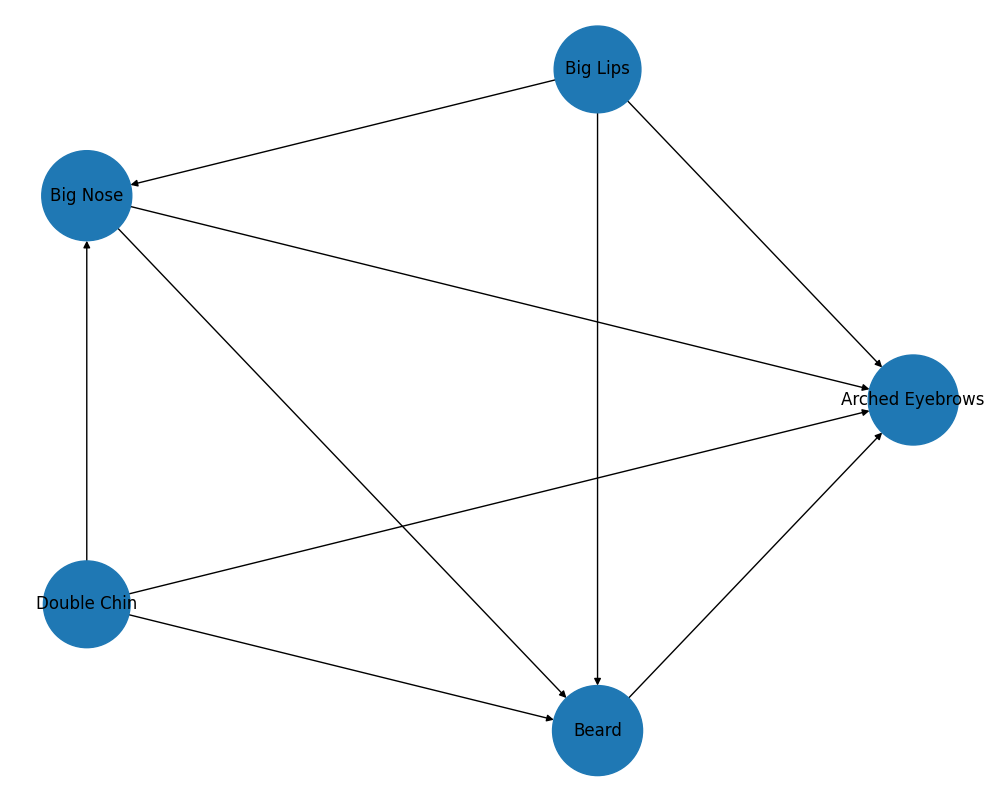}
    \caption{Bayesian Network structure learned on training set of CelebA dataset for the chosen face component attributes (\textit{big lips, arched eyebrows, big nose, double chin, beard}).}
    \label{fig:bayesian_network2}
\end{figure}

Figures~\ref{fig:bayesian_network} and~\ref{fig:bayesian_network2} illustrate the Bayesian Network structures learned from the CelebA dataset. Figure~\ref{fig:bayesian_network} presents the full network encompassing all face attributes, where the dense connectivity highlights the complex interdependencies between attributes. This complexity motivates the need for probabilistic reasoning when addressing fairness across face components, as changes to one attribute can propagate through the network.

\end{document}